\documentclass{article}

\usepackage{arxiv}

\usepackage[utf8]{inputenc} 
\usepackage[T1]{fontenc}    
\usepackage{hyperref}       
\usepackage{url}            
\usepackage{booktabs}       
\usepackage{amsfonts}       
\usepackage{amsmath} 
\usepackage{amssymb}  
\usepackage{nicefrac}       
\usepackage{microtype}      
\usepackage{cleveref}       
\usepackage{lipsum}         
\usepackage{graphicx}
\usepackage{natbib}
\usepackage{doi}
\usepackage{epsfig} 
\usepackage{mathptmx} 
\usepackage{times} 
\usepackage{bm}

\usepackage{subfig}
\usepackage{multirow}
\usepackage[export]{adjustbox}
\usepackage{url}
\usepackage[usenames]{color}
\usepackage[mathscr]{euscript}

\title{Visual Gyroscope: Combination of Deep Learning Features and Direct Alignment for Panoramic Stabilization}

\date{}

\author{ \href{https://orcid.org/0000-0003-2674-4844}{\includegraphics[scale=0.06]{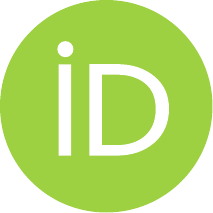}\hspace{1mm}Bruno Berenguel-Baeta}\thanks{Corresponding author.} \\
	Instituto de Investigacion en Ingenieria de Aragon\\
	Department of Computer Science and Systems Engineering\\
	University of Zaragoza\\
	Zaragoza, Spain \\
	\texttt{berenguel@unizar.es} \\
	\And
	\href{https://orcid.org/0000-0003-3318-4769}{\includegraphics[scale=0.06]{orcid.pdf}\hspace{1mm}Antoine N. Andre}\\
	CNRS-AIST JRL (Joint Robotics Laboratory)\\
	 IRL, Japan \\
 	\texttt{antoine.andre@cnrs.fr} \\
	\And
	\href{https://orcid.org/0000-0003-3703-2973}{\includegraphics[scale=0.06]{orcid.pdf}\hspace{1mm} Guillaume Caron}\\
	CNRS-AIST JRL (Joint Robotics Laboratory)\\
	IRL, Japan \\
 	Universite de Picardie Jules Verne, MIS laboratory \\
 	Amiens, France\\
 	\texttt{guillaume.caron@u-picardie.fr} \\
	\And
	\href{https://orcid.org/0000-0002-8479-1748}{\includegraphics[scale=0.06]{orcid.pdf}\hspace{1mm}Jesus Bermudez-Cameo} \\
	Instituto de Investigacion en Ingenieria de Aragon\\
	Department of Computer Science and Systems Engineering\\
	University of Zaragoza\\
	Zaragoza, Spain \\
	\texttt{bermudez@unizar.es} \\
	\And
	\href{https://orcid.org/0000-0001-5209-2267}{\includegraphics[scale=0.06]{orcid.pdf}\hspace{1mm}Jose J. Guerrero} \\
	Instituto de Investigacion en Ingenieria de Aragon\\
	Department of Computer Science and Systems Engineering\\
	University of Zaragoza\\
	Zaragoza, Spain \\
	\texttt{josechu.guerrero@unizar.es} \\
}
\renewcommand{\shorttitle}{\textit{Visual gyroscope}}
\newcommand\blfootnote[1]{%
  \begingroup
  \renewcommand\thefootnote{}\footnote{#1}%
  \addtocounter{footnote}{-1}%
  \endgroup
}
\hypersetup{
pdftitle={Visual Gyroscope: Combination of Deep Learning Features and Direct Alignment for Panoramic Stabilization},
pdfsubject={cs.CV},
pdfauthor={Bruno Berenguel-Baeta, Antoine N. Andre, Guillaume Caron, Jesus Bermudez-Cameo, Jose J. Guerrero},
pdfkeywords={Omnidirectional Vision; Image stabilization; Deep learning; Photometric algorithms},
}

\begin{document}
\maketitle
\blfootnote{A final version of this article can be found at \url{https://doi.org/10.1109/CVPRW59228.2023.00685}}

\begin{abstract}
    In this article we present a visual gyroscope based on equirectangular panoramas. We propose a new pipeline where we take advantage of combining three different methods to obtain a robust and accurate estimation of the attitude of the camera.
    We quantitatively and qualitatively validate our method on two image sequences taken with a $360^\circ$ dual-fisheye camera mounted on different aerial vehicles. 
\end{abstract}

\keywords{Omnidirectional Vision \and 3D Vision \and Non-central Cameras \and Layout recovery \and Scene understanding}

\section{Introduction}\label{sec:intro}


Autonomous navigation of Unmanned Aerial Vehicles (UAV) require the knowledge of pose and orientation. Among the different sensors we can use, cameras provide information of the surroundings of the UAV which can be used for several applications, such as object detection \cite{jiang2022review}, while being a lightweight sensor. One of these applications is the use of the camera as a Visual Gyroscope (VG), estimating the 3D orientation of the camera for a given image with respect to a reference. 

VG algorithms consider two kinds of information, direct or indirect. 
Indirect VG (IVG) leverage image features, either handcrafted, as patches around image points \cite{hadj2018closed}, or learned, \textit{e.g.} estimating first the optical flow and then the 3D orientation of the camera \cite{kim2021self}. 
Instead, Direct VG (DVG) consider pixel brightness of the whole image as input of a 3D rotation optimization method \cite{andre2022photometric}. Usually, this pixel information is transformed into different domains before estimating the orientation, such as spherical Fourier transform \cite{makadia2006rotation} or Mixture of photometric potentials \cite{caron2018spherical}.

Previous methods present high accuracy on 3D orientation estimation on narrow domains \cite{andre2022photometric}, or the possibility of a wide estimation domain but with lower accuracy \cite{makadia2006rotation}.
In this paper, we propose a hybrid pipeline combining three different methods to obtain a robust VG 
within a very large estimation domain and with high orientation accuracy. We leverage the strong points of each method, overcoming the weak points and limitations of each of them. 
In the next section we present the different methods that form our pipeline while in Sec. \ref{sec:experiments} we evaluate and compare our pipeline against state-of-the-art methods on two 360 equirectangular panoramas data-sets taken from UAV.

\section{Visual Gyroscope}\label{sec:VG}

\begin{figure*}
\centering
\includegraphics[width=0.95\textwidth]{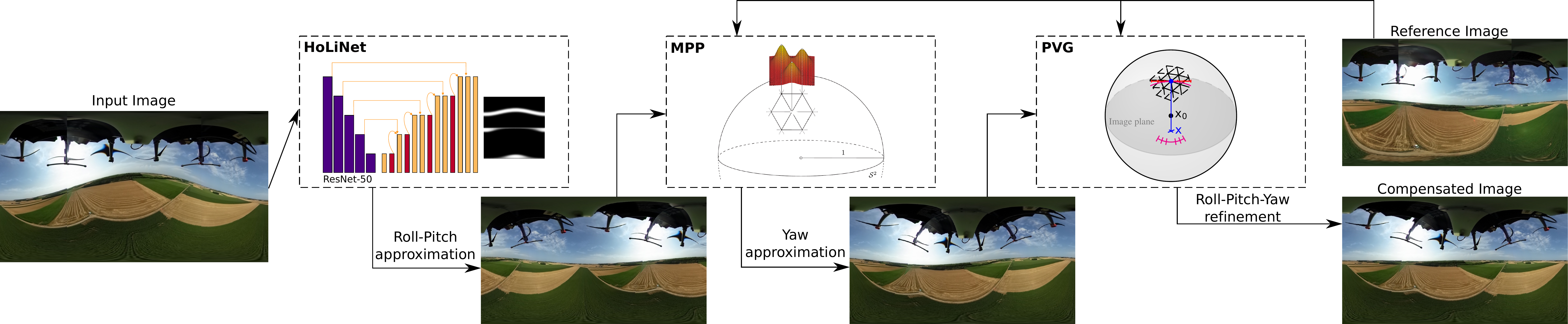}
\caption{Overview of our proposed pipeline. The input equirectangular panorama first goes through HoLiNet, obtaining a first approximation of Roll-Pitch angles with respect to the horizontal plane. The output is feed to the MPP, obtaining an approximation of the Yaw angle with respect to a reference image. Finally, the PVG refines the three angles such that the rotation compensated image is closer to the reference one.}
\label{fig:pipeline}
\vspace{-2mm}
\end{figure*}

Our proposed pipeline\footnote{Code is available in \url{https://github.com/Sbrunoberenguel/HoLiNet}} is composed by three different methods that act sequentially to obtain the Roll-Pitch-Yaw angles of the camera in outdoor environments (see Fig. \ref{fig:pipeline}). 
The first block is a IVG composed by a convolutional neural network called HoLiNet ({\textbf{Ho}rizon \textbf{Li}ne \textbf{Net}work}) where we obtain a first approximation of Roll-Pitch angles.
The second block is a DVG algorithm that uses MPP (\textbf{M}ixture of \textbf{P}hotometric \textbf{P}otentials) \cite{caron2018spherical} to retrieve the Yaw angle taking as initial solution the orientation from HoLiNet.
The third and final block is second DVG algorithm defined as PVG (\textbf{P}hotometric \textbf{V}isual \textbf{G}yroscope) where we refine the previous solutions. 

\subsection{HoLiNet}\label{subsec:holinet}

Our proposed HoLiNet follows an encoder-decoder structure taking as input information an equirectangular panorama in any orientation and providing two heat-maps, one for the horizon line and another for the vertical vanishing points (see Fig. \ref{fig:pipeline}). 

For the encoder and feature extractor we use the convolutional part of ResNet-50 \cite{he2016deep}. 
We use the available pre-trained weights, leveraging the performance of this network for feature extraction.
In the decoder we propose a set of convolutional layers followed by up-scaling layers and skip connections from the encoder part. The output of the decoder is a two-channel heat-map with the horizon line and vanishing points prediction. 
To adapt the neural network to the distortion, we use in all the convolutional layers convolutions adapted to the equirectangular projection. These convolutions, EquiConvs, were first presented in \cite{fernandez2020corners} and, in this work, are re-implemented pre-computing the kernel's distortion for a faster inference time (i.e. the original implementation runs at 0.2fps while ours can run at 10fps).

The network is trained in a collection of gravity-oriented equirectangular panoramas from indoor and outdoor environments \cite{fraguas2020}. We apply rotations to these panoramas and generate the ground truth labelling according to these rotations. We also perform several data augmentations such as: horizontal flip, color permutation, color jitter and random cropping.
At training, we obtain intermediate representations at different resolutions from the decoder of the network. These intermediate representations allow to evaluate the network on early stages, making the intermediate features of the network aim for the final task, improving the overall performance of the network.

From the output information of the network, we compute the most probable horizon plane, obtaining a first approximation of the camera's Roll-Pitch angles.
To compute the horizon plane, we use the information of both outputs of the network. 
First, we use the vanishing point heat-map to make a rough estimation of the vertical direction. We project the vanishing points heat-map in the unit sphere and compute the average vertical direction. This is a rough estimation since the output is really sparse and can be noisy.
Then, we project the horizon line heat-map into the unit sphere. Using a random sample consensus algorithm, we compute the 3D plane that better fits the horizon line heat-map in the unit sphere. 
Assuming that both network's outputs are coherent with each other, the first estimation of the vertical direction and the normal vector of the horizon plane should be close. So, to maintain this coherence, in the iterative process we discard any possible solution which differs from the first estimation more than a threshold angle (i.e. we assume a threshold of $\pm 30^\circ$). 

Once obtained the plane that better fits the output of the network, we can compute the Roll and Pitch angles of the camera (Yaw is the rotation around the plane's normal vector, which cannot be computed with this method). 
The main advantage of this method is that we can obtain 2 angles from a $360^{\circ}$ domain. The main limitation is the angle precision, which is limited by the uncertainty of the output of the network, and the impossibility to obtain the Yaw angle with the proposed representation.

\subsection{MPP-based visual gyroscope}\label{subsec:mpp}

The MPP is a full spherical image representation of image brightness as a mixture of Gaussians. The MPP is structured as an icosahedron subdivided $n \in \mathbb{N}$ times, of which each vertex is the center of a Gaussian function, weighted by the normalized intensity of the pixel where it projects in the captured image. Thus, there are as many Gaussians as vertices of the sphere but all Gaussians of the MPP share the same variance $\lambda_g \in \{ \mathbb{R} > 0 \}$ that acts as a smoothing parameter. In the seminal work introducing the MPP-based visual gyroscope~\cite{caron2018spherical}, captured dual-fisheye images are first transformed as MPPs: $G^*$ for a reference image, $G$ for a request image. Thus, $\lambda_g$ acts as a smoothing parameter of the cost function, that is the sum of squared differences between $G$ and $G^*$, to be minimized by optimizing the three angles of the 3D rotation that aligns the best both MPPs with a Newton-like algorithm. 

In this work, the differences w.r.t.~\cite{caron2018spherical} are motivated by the HoLiNet-MPP-PVG pipeline we propose (Fig.~\ref{fig:pipeline}):
\begin{itemize}
	\item \textit{input images are equirectangular} instead of dual-fisheye for both the sake of uniformity with respect to HoLiNet and more generality since the equirectangular format is much more common than dual-fisheye.
	
	\item \textit{the single Yaw angle is estimated} since not only HoLiNet already outputs estimates of Roll and Pitch angles that can serve as good initial guesses for PVG (see Sec.~\ref{subsec:pvg}) but it reduces a little the computation load.
\end{itemize}

The main advantage of the MPP-based visual gyroscope is its very large convergence domain, observed as being both globally convergent indoor for a single pure rotation angle and almost as large for 3D rotations, even in the presence of translations, in~\cite{caron2018spherical}. However, its main drawback is the required trade-off between processing time and estimation accuracy: to reach a processing rate of 3 images per second, the icosahedron maximum subdivision level is $n=3$ and the average estimation error was reported about 3 degrees with $\lambda_g = 0.325$. Thus, PVG is required as a final step to improve the accuracy of estimations. 

\subsection{Photometric Visual Gyroscope (PVG)}\label{subsec:pvg}

To improve both the processing time and the accuracy with respect to the MPP-based visual gyroscope, the PVG~\cite{andre2022photometric} uses a spherical image representation directly from image brightness without a transformation to a MPP. The PVG is built on the same principle as the MPP-based visual gyroscope and also relies on the intensities of the image mapped to a subdivided icosahedron. However, instead of computing a MPP, the direct pixel brightness are assigned to each vertices of the icosahedron, allowing a much faster computation of the Jacobians in the Newton-like optimization of the orientations. When mapping full resolution images (typically $1280\times720$) to the subdivided icosahedron, a higher precision in the orientation estimation can be obtained. The spherical gradient is computed in the image plane (as highlighted in the PVG thumbnail of Fig.~\ref{fig:pipeline}) and not with neighbors vertices as in the MPP-based gyroscope.

The main difference of this work with respect to the source work~\cite{andre2022photometric} is that \textit{equirectangular input images} are used, instead of dual-fisheye images, allowing a consistency of input images all along the orientation estimation pipeline.

PVG has shown estimation errors below $0.1^{\circ}$ in the single axis rotation case, both pure and with translations \cite{andre2022photometric}. To reach such a precision, the icosahedron is subdivided $n = 5$ times and the captured image at full resolution is used for a computation time $2\% $the one of the MPP-based gyroscope. But inversely to the MPP-based gyroscope, the low processing load and the high accuracy come at the price of a narrow convergence domain that was observed at $\pm 12.5^{\circ}$ for the single rotation angle case. Hence, PVG needs good initial guesses that HoLiNet (Roll, Pitch) and the MPP-based visual gyroscope (Yaw) bring. 

\section{Experiments}\label{sec:experiments}

\subsection{Experimental setup}\label{subsec:setup}

We evaluate our pipeline with two datasets. 
The first is the Sequence 8 of PanoraMIS~\cite{benseddik2020panoramis} that was taken with a Ricoh Theta S mounted on a Parrot Disco fixed-wing drone. Since this sequence is captured as raw dual-fisheye, we first transform the images to the equirectangular format using the software shared with the dataset\footnote{\url{http://github.com/PerceptionRobotique/dualfisheye2equi}}. With this sequence of 12000+ images, we make a qualitative evaluation by compensating the estimated rotations with respect to image 8890 in order to compare the success rate with~\cite{caron2018spherical}. 

To allow the quantitative evaluation, we introduce a new dataset of 2700+ equirectangular images, SVMIS+\footnote{\url{http://mis.u-picardie.fr/~g-caron/pub/data/SVMISplus_er_pano.zip}}. SVMIS+ is acquired with a Ricoh Theta S too but embedded on a Matrice 600 Pro hexarotor manually flown in a countryside area during $4'36"$. The use of the hexarotor has two interests over the Disco drone. First, it allows to embed an extra computer connected to the onboard electronics in order to save the orientation information of the drone's Inertial Measurement Unit synchronously with the image capture. Second, the different type of motion of a hexarotor compared to a fixed-wing drone provides an additional variety in the evaluation, 
together with different meteorological conditions 
(sunny versus cloudy).
With this dataset, in addition to the quantitative evaluation, we also perform a qualitative evaluation by compensating the estimated rotations of all the images with respect to a selected frame. Frame 1074 is selected such that the image is roughly at the center of the flown area, with attitude angles near zero (horizon line near straight and horizontal in the image). 

\subsection{Results}\label{subsec:outdoor}


With PanoraMIS' Sequence 8, we evaluate qualitatively the success rate of the computed orientations by studying the misalignment of stabilized images with respect to the reference image. Over the 12000+ images, 564 were misaligned. 
This represents a success rate of $95.3\%$, thus out-performing previously shown results of $88\%$~\cite{caron2018spherical}. The stabilization results with the whole first dataset are shared as a video\footnote{\url{http://mis.u-picardie.fr/~g-caron/videos/2023OmniCV_svmis.mp4}} highlighting each intermediate result of our pipeline for orientation estimation.

\begin{figure}
    \centering
    \includegraphics[width=0.6\textwidth]{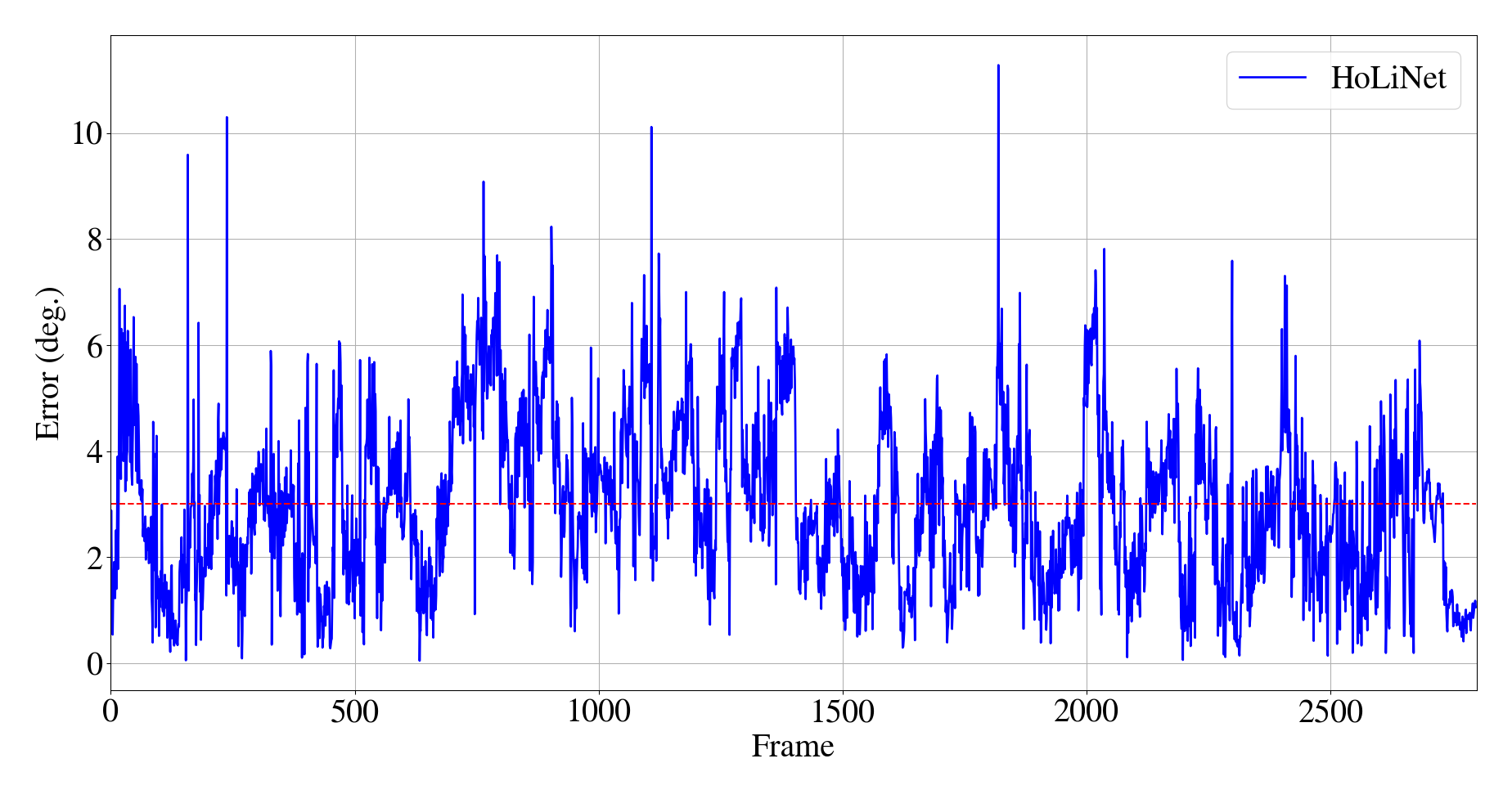}
    \caption{Quantitative results of HoLiNet for 2 angles. Dashed lines show mean error.}
    \label{fig:holineteval}
\end{figure}

With the synchronized and reliable ground truth provided with SVMIS+, a quantitative evaluation of the pipeline is put in practice.
HoLiNet is only evaluated over 2 angles. To avoid ambiguities in the evaluation of the two angles, we compute the predicted normal vector, $n$, of the horizon plane and compare with the ground truth measure, $\hat{n}$, computing the angle between both vectors as: 
$\arccos\left( n \cdot \hat{n} \right)$. The quantitative results are shown in Fig.~\ref{fig:holineteval}, obtaining a mean error of $3.04^{\circ}$. 

\begin{figure}
    \centering
    \subfloat[]{\includegraphics[width=0.48\textwidth]{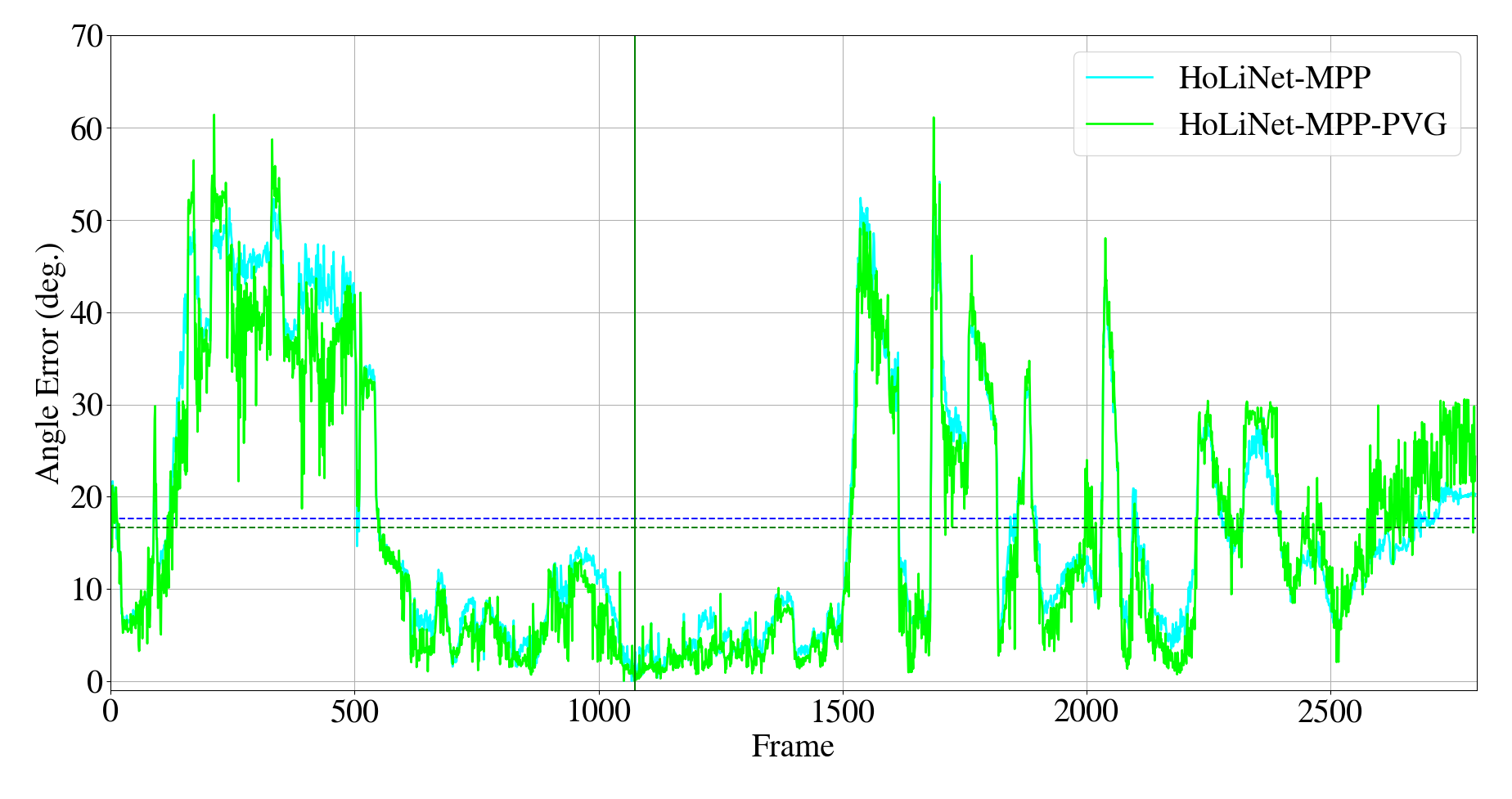}}
    \hfil
    \subfloat[\label{fig:zoom}]{\includegraphics[width=0.48\textwidth]{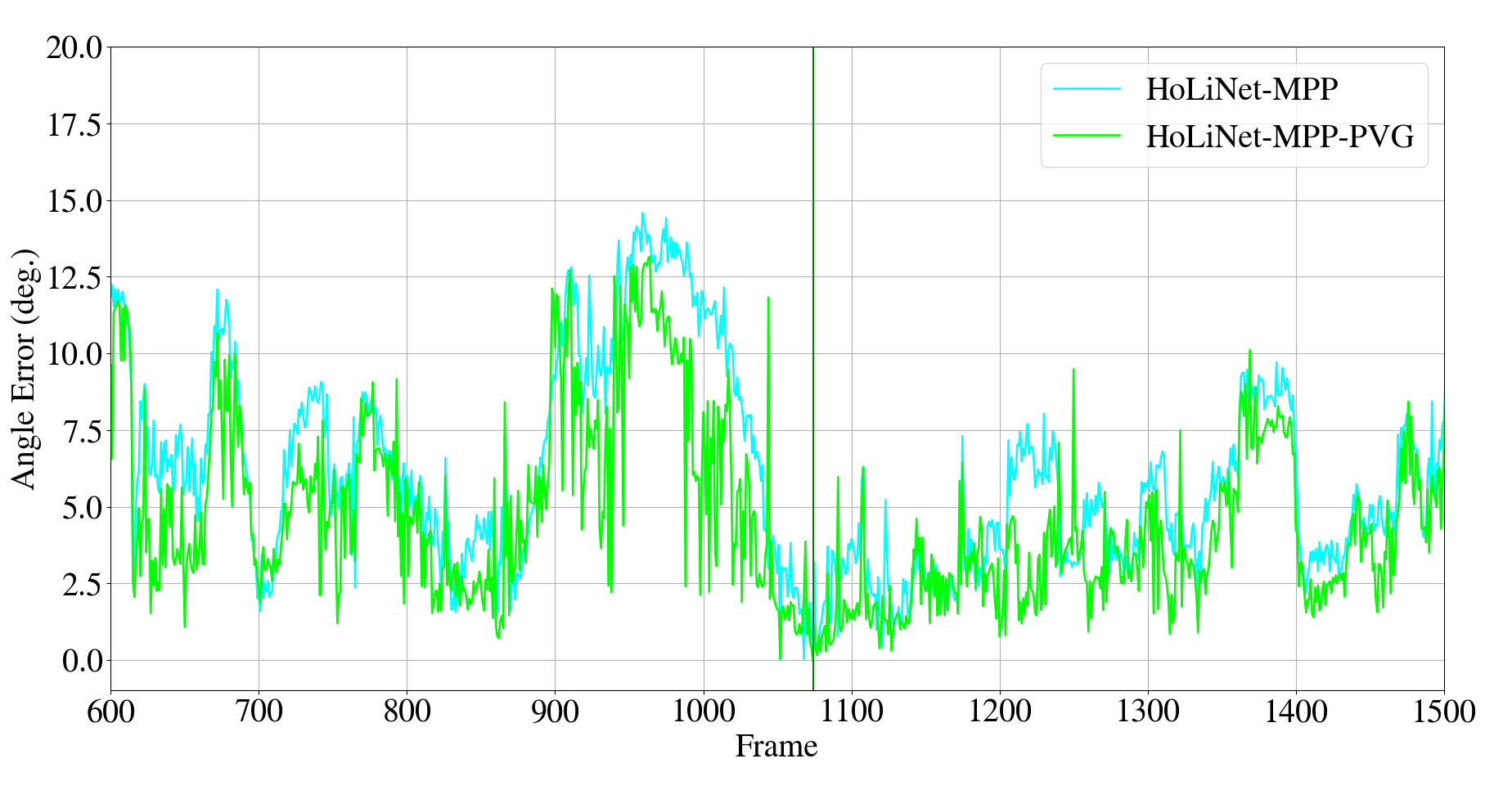}}\\
    \caption{a) Quantitative results of HoLiNet+MPP and HoLiNet+MPP+PVG in the SVMIS+ dataset. b) (Zoom) Quantitative results of MPP and PVG around the reference image. Vertical green line defines the reference image and horizontal dashed lines show mean error of each method.}
    \label{fig:evaluation}
\end{figure}

The evaluation for the 3 angles is made computing the angle difference between the 
angle-axis rotation of ground truth with respect to the reference image, $R_{gt}$, and our estimation, $R_{pred}$, as: 
$\arccos\left( (trace(R_{gt}\cdot {R_{pred}}^T) - 1) / 2\right)$.
The results are presented in Fig.~\ref{fig:evaluation}, obtaining a mean error of $17.6^{\circ}$ for HoLiNet + MPP, $16.7^{\circ}$ for the whole pipeline over the whole sequence, and $4.6^{\circ}$ for 900 images (33\% of SVMIS+) around the reference image (Fig.\ref{fig:zoom}), corresponding to a difference of altitude with respect to the reference image lower than 10~m. 
We also present a video\footnote{\url{http://mis.u-picardie.fr/~g-caron/videos/2023OmniCV_svmis_p.mp4}} of qualitative results with SVMIS+. 
\section{Conclusions}

We have presented a novel pipeline that leverages the strong points of three methods, HoLiNet using neural networks and two using direct alignment, for panoramic image rotation compensation in outdoor environments. 
The quantitative results on two datasets of airborne panoramic images show HoLiNet is robust to any translation and improves the visual alignment success rate. But since the direct alignment methods following HoLiNet in the pipeline assume pure rotation between the reference and current images, the presence of a large translation decreases the rotation estimation accuracy, as shown with the newly introduced public dataset SVMIS+. However, the rotation estimation appears accurate within a reasonable translation to the reference image that further study with SVMIS+ will allow to quantify during future works. 

\bibliographystyle{unsrtnat}
\bibliography{egbib}

\section*{ACKNOWLEDGMENT}

This work was supported by PID2021-125209OB-I00 and TED2021-129410B-I00 (MCIN/AEI/10.13039/501100011033 and FEDER/UE and NextGenerationEU/PRTR)

\end{document}